\documentclass[11pt]{article}
\usepackage[utf8]{inputenc}
\usepackage{graphicx}
\usepackage{multirow}
\usepackage{authblk}
\usepackage{hyperref}
\usepackage{geometry}
\title{\textbf{Streetscape Analysis with Generative AI (SAGAI):\\ Vision-Language Assessment and Mapping of Urban Scenes}}

\author[1]{Joan Perez\thanks{Email: \href{mailto:jperez@urbangeoanalytics.com}{jperez@urbangeoanalytics.com}, ORCID: \href{https://orcid.org/0000-0003-3003-0895}{0000-0003-3003-0895}}}
\author[2]{Giovanni Fusco\thanks{Email: \href{mailto:giovanni.fusco@univ-cotedazur.fr}{giovanni.fusco@univ-cotedazur.fr}, ORCID: \href{https://orcid.org/0000-0002-6171-5486}{0000-0002-6171-5486}}}

\affil[1]{Urban Geo Analytics, France}
\affil[2]{Université Côte-Azur-CNRS-AMU-Avignon Université, ESPACE, France}

\date{April 2025}

\begin{document}

\maketitle

\begin{abstract}
Streetscapes are an essential component of urban space. Their assessment is presently either limited to morphometric properties of their mass skeleton or requires labor-intensive qualitative evaluations of visually perceived qualities. This paper introduces SAGAI: Streetscape Analysis with Generative Artificial Intelligence, a modular workflow for scoring street-level urban scenes using open-access data and vision-language models. SAGAI integrates OpenStreetMap geometries, Google Street View imagery, and a lightweight version of the LLaVA model to generate structured spatial indicators from images via customizable natural language prompts. The pipeline includes an automated mapping module that aggregates visual scores at both the point and street levels, enabling direct cartographic interpretation. It operates without task-specific training or proprietary software dependencies, supporting scalable and interpretable analysis of urban environments. Two exploratory case studies in Nice and Vienna illustrate SAGAI’s capacity to produce geospatial outputs from vision-language inference. The initial results show strong performance for binary urban–rural scene classification, moderate precision in commercial feature detection, and lower estimates, but still informative, of sidewalk width. Fully deployable by any user, SAGAI can be easily adapted to a wide range of urban research themes, such as walkability, safety, or urban design, through prompt modification alone.
\end{abstract}

\vspace{1em}

\section{Introduction}
Assessing the qualities and functions of urban streetscapes is essential to understand walkability, safety, commercial vitality, and social life in cities \cite{gehl2013, harvey2016, dovey2018}. However, traditional methods for observing and evaluating street-level conditions, such as field surveys, audits, and manual photo interpretation, remain time-consuming, labor-intensive, and difficult to scale beyond small pilot zones \cite{harvey2016}. Geo-processing of vector models of the built environment allows the assessment of morphometric properties of the skeletal streetscape \cite{harvey2017, araldi2025}, that is, of the 3D masses defining the visual channel of the streetscape. However, more fine-grained aspects of the streetscape, including building façades, urban furniture, sidewalks, materials and texture of built-up elements, greenery, cleanliness, etc. can play a fundamental role for human perception and usage of the public space of the street. These elements are collectively referred to as the skin of the streetscape \cite{harvey2017}, and are rarely the object of automated assessment through street imagery.   Clarke et al. \cite{clarke2011} and Rundle et al. \cite{rundle2011} are among the first to propose the use of streetscape images accessed online to produce a complete assessment of the streetscape, but they end up using traditional manual audit methods remotely instead of on site. Although convolutional neural networks (CNNs) and object detection models have later enabled partial automation through supervised classification of images, these systems typically require extensive labeled datasets and task-specific training pipelines, making them costly to deploy at scale and limited in scope to narrowly defined features or categories.

Recent advances in generative vision-language models (VLMs) offer a promising alternative. By combining image understanding with natural language prompting, models like LLaVA allow structured information to be extracted directly from visual data, enabling rapid and automated interpretation of complex street-level scenes. This opens new opportunities for scaling urban analysis, especially when paired with the growing availability of georeferenced street imagery platforms such as Google Street View.

The present research develops and operationalizes SAGAI (Streetscape Analysis with Generative Artificial Intelligence), a four-step workflow designed to automate the application of such protocols using open-access geospatial data and vision-language models. SAGAI enables reproducible and scalable image-based scoring on large urban areas. Built entirely in Python and deployable in free-tier Google Colab environments, the workflow links OpenStreetMap, Google Street View, and LLaVA to generate structured indicators of the built environment at street scale. Each of the four modules—point generation, image retrieval, model inference, and spatial aggregation—is designed to be modular, transparent, and adaptable to a wide range of urban research questions.

The remainder of this paper is organized as follows. Section ~\ref{sec:2} reviews related work in streetscape analysis, AI applications in urban contexts, and recent developments in vision-language models. Section ~\ref{sec:3} presents the SAGAI workflow, describing each of its four modules. Section ~\ref{sec:4} illustrates the pipeline through two case studies in Nice, France and Vienna, Austria. Section ~\ref{sec:5} reports accuracy results from manual validation and highlights interpretative gaps between AI predictions and human annotations. Section ~\ref{sec:6} discusses performance, limitations, and future extensions. We conclude in Section ~\ref{sec:7} by outlining implications for urban analysis and open-source tool development.
\section{Related Work}
\label{sec:2}
Research on streetscape analysis has long relied on computer vision to extract information from urban imagery, particularly from platforms like Google Street View. Before the rise of generative artificial intelligence (GenAI), urban studies predominantly employed discriminative models—algorithms designed to detect and classify specific visual features using pre-labeled datasets. A foundational example is Treepedia \cite{li2018}, which calculated a Green View Index through semantic segmentation of GSV panoramas to quantify urban tree canopy. Street-level imagery has since become a vital data source in urban research, enabling a range of applications: detecting sidewalks \cite{hosseini2022}, measuring visual clutter from vegetation \cite{seiferling2017}, classifying façade-level attributes such as fences and setbacks \cite{law2018}, and even inferring neighborhood demographics from the types of visible vehicles \cite{gebru2017}. Recent works such as Urban Visual Intelligence \cite{zhang2024} consolidate these approaches into a structured hierarchy, leveraging discriminative AI and deep learning techniques—such as semantic segmentation and scene classification—for large-scale physical and socioeconomic analysis of cities based on street-level images.

These studies typically relied on convolutional neural networks (CNNs) and object detectors such as R-CNN \cite{girshick2014} or YOLO \cite{redmon2016}, trained on extensive labeled corpora via supervised or semi-supervised learning. While effective, such models are constrained by their architecture and training scope—optimized for specific tasks, rather than for contextual interpretation or semantic abstraction. Their generalizability depends heavily on dataset coverage and annotation quality, often requiring domain-specific training pipelines and hard-coded feature engineering.

Parallel to this, the field of geospatial artificial intelligence (GeoAI) is emerging as a broader paradigm integrating spatial data, machine learning, and advanced computational tools. GeoAI applications have proliferated across multiple domains of human geography—including urban morphology, mobility, health, and inequality—leveraging spatial imagery (satellite, aerial, or street-level) for classification, prediction, and simulation tasks \cite{wang2024}. Recent work further demonstrates GeoAI’s capacity to map building footprints, classify land use, and model natural hazards using multimodal datasets \cite{song2023}. However, despite these advancements, most GeoAI pipelines remain discriminative in nature—focused on identifying what is in a given image, rather than generating data, novel descriptions or simulating transformations from a given image.

A shift is currently underway with the emergence of vision-language models (VLMs)—a class of generative architectures that jointly process images and text within a unified framework. Unlike traditional discriminative models trained on object labels and bounding boxes, VLMs learn from large datasets of image–text pairs, enabling more flexible and contextual interpretation of visual scenes. These models typically consist of two core components: a vision encoder, which extracts visual features from images, and a large language model (LLM), which handles the textual input, performs reasoning, and generates responses.

Models such as LLaVA \cite{liu2023}, BLIP-2 \cite{li2023}, GPT-4 with Vision \cite{openai2023}, and Flamingo \cite{alayrac2022} all follow this general architecture. A widely adopted vision encoder in these systems is CLIP \cite{radford2021}, an open-source model trained to align image and text representations in a shared embedding space via contrastive learning. While CLIP is commonly used due to its performance and accessibility, other encoders—such as Vision Transformers (ViT) \cite{dosovitskiy2020} or Swin Transformers \cite{liu2021} —can also fulfill this role depending on the model.

In the case of LLaVA, the architecture explicitly wraps these components with an intermediate projection layer that aligns the image embeddings from the vision encoder with the token embeddings expected by the LLM. This enables multimodal inference, such as answering image-based questions, generating descriptions, or reasoning about spatial configurations, without task-specific retraining. Its open-source nature and efficient implementation—including quantized variants for deployment on consumer-grade hardware—make LLaVA especially suited for research and practical applications in geospatial and urban analytics.

Despite these advances, the potential of vision-language models (VLMs) for geospatial and urban research remains largely underexplored. A few recent studies have begun to experiment with approaches that resonate with the principles underlying SAGAI. For example, StreetViewLLM integrates street-level imagery with geographic metadata to infer location and semantic context, employing chain-of-thought prompting to enhance spatial reasoning \cite{li2024}. Similarly, CityLLaVA adapts a LLaVA-style architecture to urban safety scenarios, combining structured prompts and bounding box inputs to interpret city environments \cite{duan2024}. Blečić et al. \cite{blecic2024} propose a multimodal LLM-based workflow for assessing urban walkability from street view images, which augments visual scoring with textual interpretation to support planning decisions. In a complementary direction, Wei et al. \cite{wei2024} introduce GeoTool-GPT, a fine-tuned LLaMA-2 model designed to interface with GIS tools, while Kazemi Beydokhti et al. \cite{kazemi2024} explore qualitative spatial reasoning for geospatial question answering. Additionally, Schmidt et al. \cite{schmidt2025} evaluate the spatial accuracy of several VLMs—including LLaVA1.6 and GPT-4o—in geocoding flood-related imagery.

Although none of these studies offer fully integrated geospatial analysis pipelines, they reflect a growing interest in applying large multimodal and language models to urban domains. SAGAI contributes to this evolving field by combining generative vision-language reasoning with reproducible geospatial workflows, explicitly linking semantic image interpretation with open-access spatial datasets. SAGAI was developed within the emc2 research project \cite{fusco2024}, aiming to assess the potential of suburban areas for a human-centred 15-minute city model. Within the project, urban patterns were identified within streetscapes of test areas, combining fieldwork data and ad hoc geoprocessing tools \cite{fusco2025}. The need emerged to automate the workflow using street imagery as input and performing combined operations of object recognition, classification, counting and measurement to assess the presence/absence of the selected patterns, above all at the level of the streetscape skin. SAGAI is thus the first step towards automating complex operations of streetscape assessment. It is designed to generate interpretable numeric indicators—such as walkability, conviviality or safety scores—at scale, while preserving sensitivity to urban street network topology and human-centred planning theory \cite{alexander1977, gehl2010}. By doing so, SAGAI bridges the gap between AI-powered image analysis and actionable urban insights, enabling customizable prompts and spatialized outputs across diverse urban themes.
\section{Methodology: The SAGAI Workflow}
\label{sec:3}
SAGAI is structured into four modules, each implemented in Python and designed to run in Google Colab. All scripts use Google Drive for reading inputs and saving outputs, requiring a one-time authentication step at the beginning of each session. The four modules are described below.
\begin{figure}[htbp]
  \centering
  \includegraphics[width=0.85\textwidth]{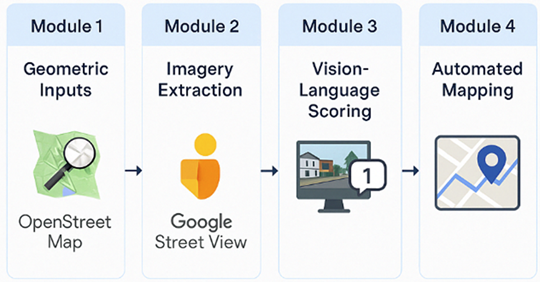}
  \caption{The Four Modules of the SAGAI Workflow}
  \label{fig1}
\end{figure}
\subsection{Module 1: OSM Point Generator}
The first module of the SAGAI workflow is designed to generate sample points along the street network of a user-defined area. It operates without requiring any input data beyond a bounding box, as it automatically retrieves the relevant street geometries from OpenStreetMap. This ensures global applicability while minimizing the preparation burden on the user.

Once the network is downloaded, points are distributed along the streets based on two user-defined parameters: spacing and offset. The spacing determines the distance between points, while the offset prevents placing points too close to intersections. Each point is associated with its corresponding street segment via a unique identifier, enabling subsequent spatial aggregation steps.

The output consists of a single GeoPackage containing two layers: the cleaned street network and the set of generated points. These are projected to appropriate coordinate systems for accurate distance calculations and then reprojected to geographic coordinates for compatibility with later modules. This general-purpose module provides the spatial input for image retrieval and visual scoring tasks.
\subsection{Module 2: Street View Batch Downloader}
The second module of SAGAI automates the retrieval of street-level imagery from Google Street View based on the geographic points generated in the first module. For each location, the script sends requests to the Google Street View Static API to download images in four predefined directions—typically the cardinal angles (0°, 90°, 180°, and 270°). The number and orientation of views can be adjusted by the user, as can camera parameters such as pitch and field of view angles. To detect missing or invalid imagery, a filtering step is applied immediately after download: images dominated by a single pixel value—commonly used in Google’s “no imagery” placeholders—are flagged accordingly.

The user must provide an active API key with access to the Street View Static API. The output is a structured dataset of georeferenced images that serves as input for the subsequent image-based scoring module. Designed for high-throughput execution in the Google Colab environment, the module can scale efficiently to thousands of locations with minimal user intervention.
\subsection{Module 3: Scene Assessment with LLaVA}
The third module is the core of the SAGAI pipeline. It performs automated image-based scoring using the LLaVA v1.6 model with a Mistral-7B backbone. It applies structured prompts to each image to extract specific visual information—such as classification as urban scene, commercial presence, or sidewalk width—and returns a corresponding numerical score. The model is loaded in 4-bit quantized format to support efficient inference in memory-constrained environments. All components are open-access and retrieved from Hugging Face, a widely used repository for distributing pretrained machine learning models. By default, the script uses the publicly available checkpoint (a snapshot of the model's state at a particular point during training) liuhaotian/llava-v1.6-mistral-7b, which can be downloaded without authentication. If users wish to use private weights, a Hugging Face account and access token may be required.

A custom scoring function is implemented to structure the interaction between the image input and the LLaVA model. It follows LLaVA’s conversation architecture, embedding the prompt with image tokens and formatting it using role-based templates. Images are preprocessed using CLIP-compatible vision encoders and inference is performed with low-temperature sampling and stopping criteria adapted to extract concise scalar outputs. Temperature sampling adjusts the randomness of the model’s predictions—lower values make the output more deterministic, which is desirable for scoring tasks. Stopping criteria are used to terminate the generation process once a specific token or pattern is detected, ensuring that only the relevant part of the response is returned. This controlled generation strategy helps enforce the expected numerical format and improves reproducibility across sessions.

Each image is processed independently using a user-defined prompt that encodes a scoring rule in natural language. The default configuration includes three pre-coded scoring tasks: (i) binary classification of urban versus rural scenes, (ii) shopfront counting, and (iii) estimation of sidewalk width. The desired task is selected through a task identifier (T1, T2, or T3), which activates the corresponding prompt template. These prompts can be freely modified or replaced by the user to implement other visual feature scoring strategies.

The script automatically reads the images retrieved by Module 2. Files previously marked as unavailable are skipped from visual analysis but still recorded in the output table to preserve traceability. A resuming mechanism allows the script to continue from the last processed image if an output file already exists. This feature is particularly useful when working with large-scale study areas, such as entire metropolitan regions, where long runtimes or interruptions are common.

The module is fully automated and requires minimal user input. The numerical scores produced in this module serve as direct inputs for the final stage of the pipeline, which aggregates and maps the results at the spatial level.
\subsection{Module 4: Geospatial Scoring Aggregation and Mapping}

The fourth module of the SAGAI workflow aggregates the image-based scores and integrates them into a geospatial dataset. It takes as input the numerical outputs generated in Module 3 along with the spatial geometries produced in Module 1, and summarizes the values at both the point and street segment levels.

All four summary statistics are retained in the output GeoPackage for both points and streets. The user can then select either the mean or sum values for cartographic rendering, depending on the task and desired interpretation. Indeed, a visualization tool is integrated to render two thematic maps: one showing the values assigned at the point level and the other the scores at the street level. Streets and points without any valid data are displayed in grey. 

The module requires three inputs: the case study name, the task identifier, and the aggregation mode for the maps. This final step of the pipeline allows users to translate raw visual assessments into interpretable spatial scores suitable for urban analysis, cartographic representation, or downstream statistical modeling. The outputs can serve as a foundation for further spatial analysis. In particular, the point-level indicators can be passed into complementary methods that retain the underlying street-based topology (e.g. ILINCS \cite{yamada2010}) or to identify recurring spatial configurations.
\subsection{Open-Source Availability and Repository Access}
\label{sec:github}
The full implementation of SAGAI (v1.0) is released as open-source software and publicly accessible at:

\url{https://github.com/perezjoan/SAGAI}

The repository includes:
\begin{itemize}
    \item Modular Python scripts for all four components of the pipeline;
    \item Predefined prompts for the three scoring tasks (urban classification, storefront counting, and sidewalk width estimation);
    \item Sample inputs and outputs for the Nice and Vienna case studies;
    \item Colab-ready notebooks enabling direct execution in the cloud;
    \item Detailed “NOTICE” files for each module, covering inputs, configuration, and expected outputs.
\end{itemize}

The platform is designed for zero-shot deployment on free-tier Google Colab environments. Only minimal configuration (bounding box coordinates, API key, and task identifier) is required to run the full pipeline. No local installation or specialized hardware is needed. The codebase is released under an open license; details on dependencies and usage rights are provided in the repository’s LICENSE file.

\section{Empirical Evaluation: Two Case Studies in the Peripheral Areas of Nice and Vienna}
\label{sec:4}
To illustrate the capabilities of the SAGAI workflow, two contrasting urban environments were selected: the Paillon Valley in the northeastern outskirts of Nice, France, and the western districts of Penzing and Wolfersberg in Vienna, Austria. Both areas present complex spatial structures but differ significantly in morphology, density, and land use. The Paillon Valley follows a narrow strip of flatland, linking the housing project of L’Ariane, in Nice, with the surrounding municipalities. It is characterized by linear urbanization, mixed land uses, and constrained topography. In contrast, the Penzing-Wolfersberg sector of Vienna lies at the foothills of the Wienerwald and features a low-density residential fabric with allotment gardens, winding roads, and forested patches. These case studies were chosen for their heterogeneity and the diversity of visual environments they offer, enabling the evaluation of SAGAI across distinct urban forms.
\begin{figure}[htbp]
  \centering
  \includegraphics[width=0.85\textwidth]{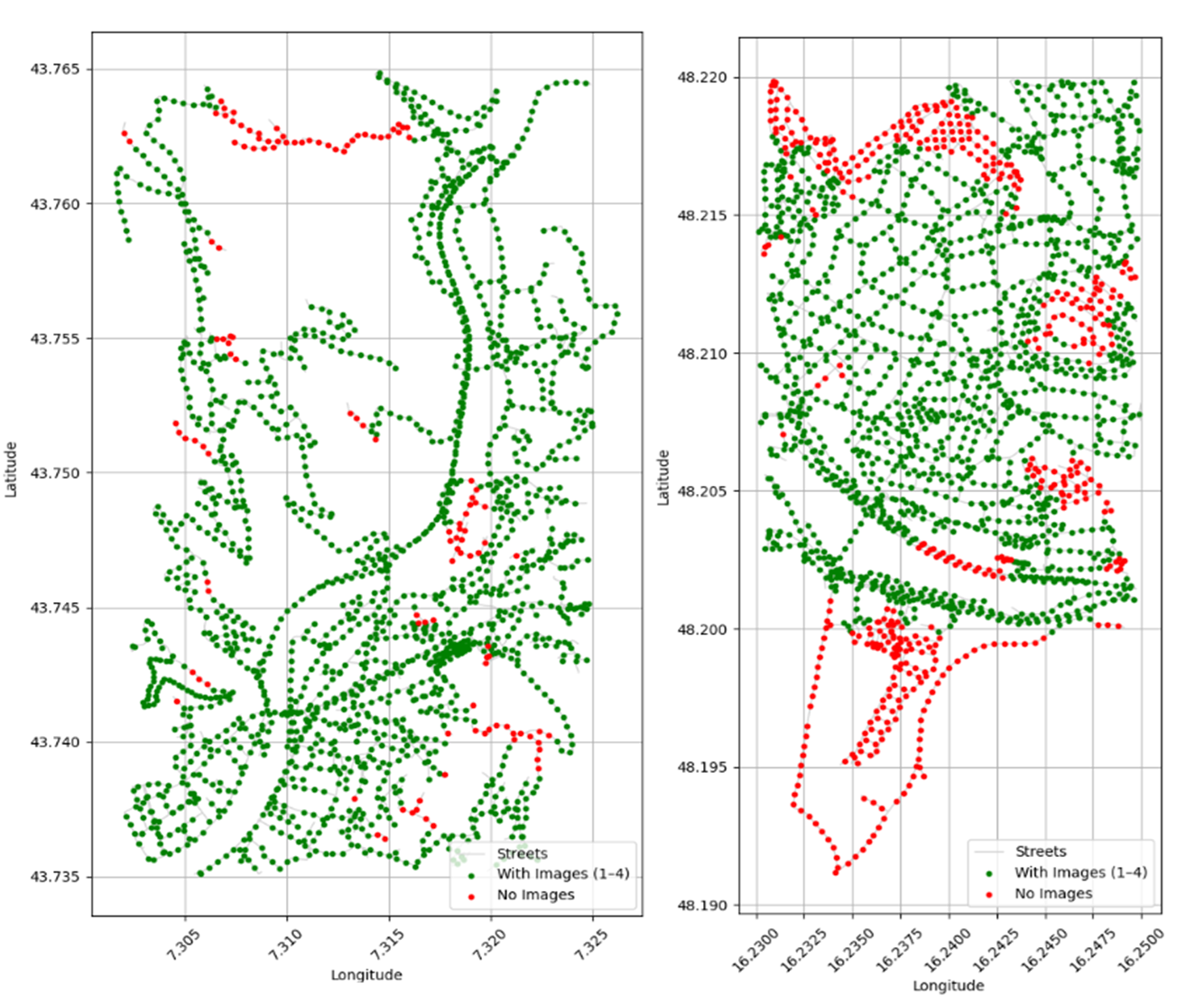}
  \caption{Street View Image Availability by Sample Point in Nice (left) and Vienna (right)}
  \label{fig2}
\end{figure}
Figure ~\ref{fig2} presents the spatial distribution of sampled points and their corresponding Street View image availability for the two case studies — Paillon Valley (Nice) on the left and Penzing–Wolfersberg (Vienna) on the right. In both areas, points were generated during module 1 along the OSM street network using a fixed offset of 15 meters from intersections and a spacing interval of 40 meters. The tabulated metrics show that despite covering a smaller bounding box (4.96 km²), the Vienna case includes more street segments (1228 vs. 955) and a slightly longer total street length (154.84 km vs. 141.31 km), reflecting the denser layout of the street network in the area.

The Nice case study achieved near-complete coverage, with 1775 out of 1898 points successfully associated with four valid images. Only 123 points (6.5\%) lacked coverage. In contrast, the one in Vienna shows significant coverage gaps: 449 of 1948 points (23\%) returned no imagery. These gaps, visualized as red points in the figure, are especially concentrated along the southern and northeastern fringes of the Vienna study area. These discrepancies are not the result of the sampling procedure but rather reflect underlying differences in Street View coverage. The lack of imagery is often associated with areas that are poorly served by vehicular access, such as cul-de-sacs, pedestrian-only paths, or segments located within forested or semi-private residential zones. These features are more prevalent in the hilly and low-density morphology of Penzing and Wolfersberg, explaining the higher rate of missing images relative to the more continuous and linear urban form of Paillon Valley in Nice.
\begin{table}[htbp]
\centering
\begin{tabular}{|l|c|c|}
\hline
\textbf{Metric} & \textbf{Nice (Paillon Valley)} & \textbf{Vienna (Penzing \& Wolfersberg)} \\
\hline
Number of street segments & 955 & 1228 \\
Total street length & 141.31 km & 154.84 km \\
Total number of points & 1898 & 1948 \\
Bounding box surface & 6.18 km\textsuperscript{2} & 4.96 km\textsuperscript{2} \\
Points with 4 images & 1775 & 1499 \\
Points with no coverage & 123 & 449 \\
\hline
\end{tabular}
\caption{Street Network and Imagery Statistics for the Nice and Vienna Case Studies}
\label{tab:coverage}
\end{table}

To evaluate the SAGAI workflow, all three predefined tasks of module 3 are deployed in both case studies. These include: (T1) a categorization task to classify scenes as either urban or rural; (T2) a counting task to detect the number of visible commercial storefronts; and (T3) a measuring task to estimate the visible width of sidewalks. Each task is implemented using a modular prompt architecture that guides the LLaVA vision-language model to return concise scalar values. Prompts are dynamically assembled from four components \textit{— \{role\_description\}, \{theory\_model\}, \{task\},} and \textit{\{response\_format\}} — allowing for flexible adaptation to different visual features or scoring schemes. The resulting outputs range from binary classifications (T1), to cardinal counts (T2), and continuous numeric estimates (T3). This triad of tasks demonstrates SAGAI’s capacity to extract diverse visual information all relevant to the assessment of urban streetscapes —establishing a robust and extensible approach for comparative spatial analysis of streetscapes. Complete prompt examples used in version 1.0 of SAGAI are included in \hyperref[app:prompts]{Appendix~A}.

The scoring outputs generated for each image are then aggregated spatially following the procedure described in Module 4. For every sampled point, the scores obtained across four directional images are summarized into a single point-level metric by averaging and summing valid values. These point-level results are further linked to their corresponding street segments to produce aggregated scores at the street level. The following section presents the results of this geospatial aggregation for each task, comparing distributions across the Nice and Vienna study areas.
\section{Experimental Results}
\label{sec:5}
To evaluate the reliability of SAGAI, a manual validation procedure was conducted across the three predefined scoring tasks—categorization (T1), counting (T2), and measuring (T3)—in both case study areas. A stratified random sample of 300 image-level predictions was selected and compared against human annotations. For each task, class-specific precision and total accuracy were calculated. Ambiguous cases—where the human annotator could not confidently assign a score due to unclear visual content, image truncation, or corrupted input—are marked as “NA”. These are excluded from accuracy calculations but still counted in the reported sample sizes.

Table~\ref{tab:accuracy} summarizes the precision and overall accuracy results for each city and task. In Task 1 (urban vs. rural categorization), where the model assigns a score of 0 for rural and 1 for urban, SAGAI achieved its highest performance, with 92.73\% total accuracy. In Task 2 (storefront counting), which uses 0 for no visible storefronts, 1 for one storefront, and 2+ for multiple, performance declined—particularly for intermediate categories. Task 3 (sidewalk width estimation), which accepts a range of values such as 0, 0.5, 1.0, 1.5, and so on, showed the lowest precision overall. This reflects the increased complexity and difficult objectification of fine-grained visual assessments.

\renewcommand{\arraystretch}{1.2} 
\setlength{\tabcolsep}{6pt} 

\begin{table}[htbp]
\centering
\resizebox{\textwidth}{!}{%
\begin{tabular}{|l|l|c|c|c|c|c|c|c|}
\hline
\textbf{Task} & \textbf{Location} & \textbf{Precision 0} & \textbf{Precision 0.5} & \textbf{Precision 1} & \textbf{Precision 1.5} & \textbf{Precision 2+} & \textbf{Overall Accuracy} & \textbf{NA Cases} \\
\hline
\multirow{3}{*}{\textbf{T1:\ Categorization}} 
& Nice   & 95.83\% (23/24) & — & 84.00\% (21/25) & — & — & 89.80\% (45/50) & 1 \\
& Vienna & 91.30\% (21/23) & — & 91.67\% (22/24) & — & — & 91.49\% (43/47) & 3 \\
& Total  & —               & — & —               & — & — & 91.67\% (88/96) & 4 \\
\hline
\multirow{3}{*}{\textbf{T2:\ Counting}} 
& Nice   & 90.00\% (18/20) & — & 70.00\% (14/20) & — & 60.00\% (12/20) & 73.33\% (44/60) & 0 \\
& Vienna & 100.00\% (20/20) & — & 45.00\% (9/20) & — & 20.00\% (4/20) & 55.00\% (33/60) & 0 \\
& Total  & —               & — & —               & — & —              & 64.17\% (77/120) & 0 \\
\hline
\multirow{3}{*}{\textbf{T3:\ Measuring}} 
& Nice   & 71.43\% (10/14) & 25.00\% (3/12) & 53.85\% (7/13) & 53.33\% (8/15) & — & 51.85\% (28/54) & 6 \\
& Vienna & 53.33\% (8/15) & 61.54\% (8/13) & 64.29\% (9/14) & 46.67\% (7/15) & — & 56.14\% (32/57) & 3 \\
& Total  & —              & —              & —              & —              & — & 54.05\% (60/111) & 9 \\
\hline
\end{tabular}
}
\caption{Accuracy and Class-Specific Precision for Each Scoring Task in Nice and Vienna (Random Samples, Human Validation)}
\label{tab:accuracy}
\end{table}

\begin{figure}[htbp]
  \centering
  \includegraphics[width=0.85\textwidth]{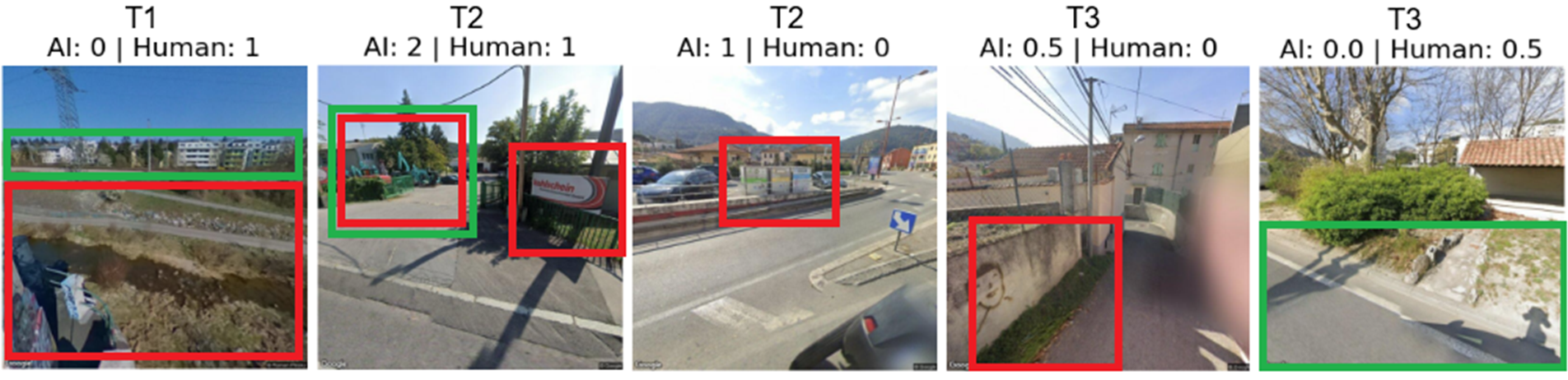}
  \caption{Examples of Misclassified Images with Highlighted Interpretation Gaps between Human and AI Scoring}
  \label{fig3}
\end{figure}

Beyond the overall accuracy metrics, manual inspection revealed systematic differences in how the model interprets visual scenes compared to human annotators. First, we noted that ambiguity is not exclusive to AI since certain cases proved challenging even for human annotators. Figure ~\ref{fig3} illustrates several examples across the three tasks (T1–T3) where the model's predictions diverged from human judgment. Red boxes\footnote{Red and green rectangles are not object detection bounding boxes, but visual annotations added manually to highlight interpretation differences between human and AI assessments.} illustrate features that were likely misinterpreted by the AI while the green boxes highlight the elements that human annotators considered relevant when assigning their score.

For Task 1 (urban vs. rural classification), failures typically stemmed from hybrid landscapes where natural and built elements coexist. In the example shown, the model misclassified a clearly urban setting as rural, possibly due to the presence of large green spaces in the foreground. In Task 2 (storefront counting), we observed common confusions with non-commercial elements or with commercial elements which are not storefronts. The model often interpreted advertising signs, parked commercial vehicles, utility structures or clusters of waste containers as storefronts. Task 3 (sidewalk width estimation) presented the most diverse range of misinterpretations. One common issue was the consistent identification of grass strips—typical of North American residential streets—as sidewalks, even when they were clearly not usable pedestrian space. Painted edge lines serving as pedestrian sidewalks were also often not detected as sidewalks by the AI. 

Another recurring issue arose when two sidewalks appeared in the same image: without instruction on which side to evaluate, the model made inconsistent or averaged estimations. While this ambiguity had limited impact when sidewalks were of similar widths on both sides, it remained a source of error in other configurations. Notably, the model never returned a width above 2 meters, even when wider sidewalks were visible in the validation sample. Yet, not all mismatches should be interpreted as complete failures. For instance, detecting one storefront instead of two, or slightly underestimating a sidewalk width, is qualitatively better than returning no detection at all. That said, some failure patterns can’t be explained. For example, in certain images with no visible sidewalk at all, the model still returned a non-zero width estimate.

These examples illustrate how the LLaVA v1.6 Mistral-7B model, despite performing well in structured tasks, struggles with contextual reasoning in more complex urban scenes. They also highlight the value of integrating options for expressing uncertainty in the prompts. For the full set of annotated examples and corresponding scores, readers are referred to \hyperref[app:validation]{Appendix~B}.

\begin{figure}[htbp]
  \centering
  \includegraphics[width=1\textwidth]{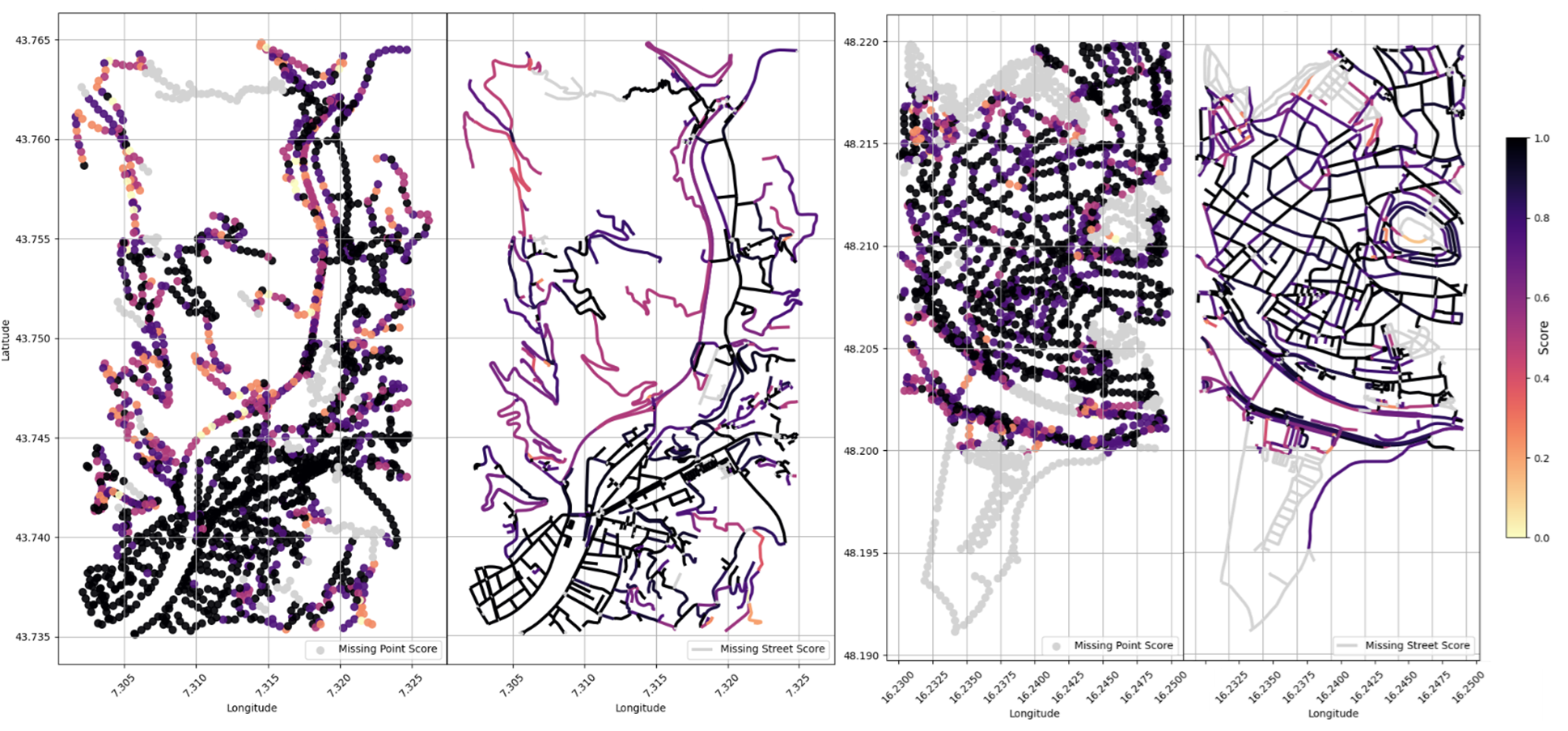}
  \caption{Averaged Urban Character Scores from Module 4 – Task 1 (Categorization) for Nice (left) and Vienna (right), at Point and Street Segment Levels}
  \label{fig4}
\end{figure}
Figure ~\ref{fig4} shows the spatial distribution of predicted urban scores from Module 4 – Task 1 (categorization) for the two case study areas: Nice (left) and Vienna (right). The left panel of each city map shows averaged point-based predictions, while the right panel displays averaged scores at the street segment level. Indeed, even point-based predictions have average values of the urban character of the scene, as 4 street-view images are normally associated with each point. A continuous color scale from light yellow (0: rural) to dark purple (1: urban) reflects the model’s results, with grey elements representing locations where no prediction could be made due to missing imagery.

In both cities, the model clearly differentiates central urban cores from peripheral or more natural areas. In Nice, the densely urbanized housing project of L’Ariane and the valley floor stand out with high scores, while hillside streets and isolated paths along the Paillon River tend to receive lower urban character values. Similarly, in Vienna, the dense road network of Penzing is consistently classified as urban, while peripheral allotment zones and forest-adjacent roads receive lower scores. The aggregated street-level visualizations help reveal broader structural patterns and remove isolated point-level noise, highlighting the value of spatial smoothing.
\begin{figure}[htbp]
  \centering
  \includegraphics[width=0.9\textwidth]{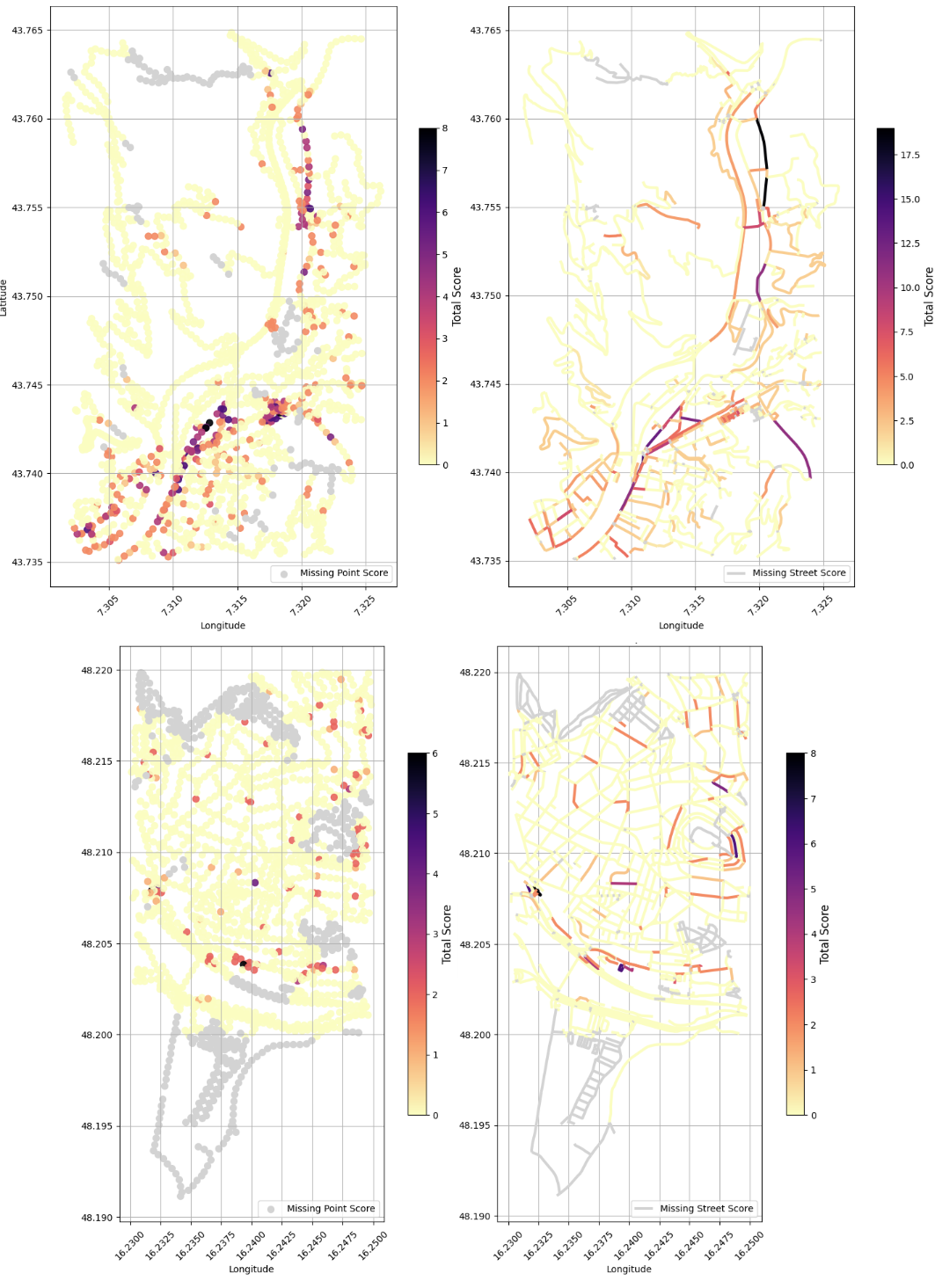}
  \caption{Total predicted scores for storefront presence at the point level (left subpanels) and aggregated by street segment (right subpanels) from Module 4  – Task 2 (Counting) shown for Nice (top) and Vienna (bottom)}
  \label{fig5}
\end{figure}

\begin{figure}[htbp]
  \centering
  \includegraphics[width=0.9\textwidth]{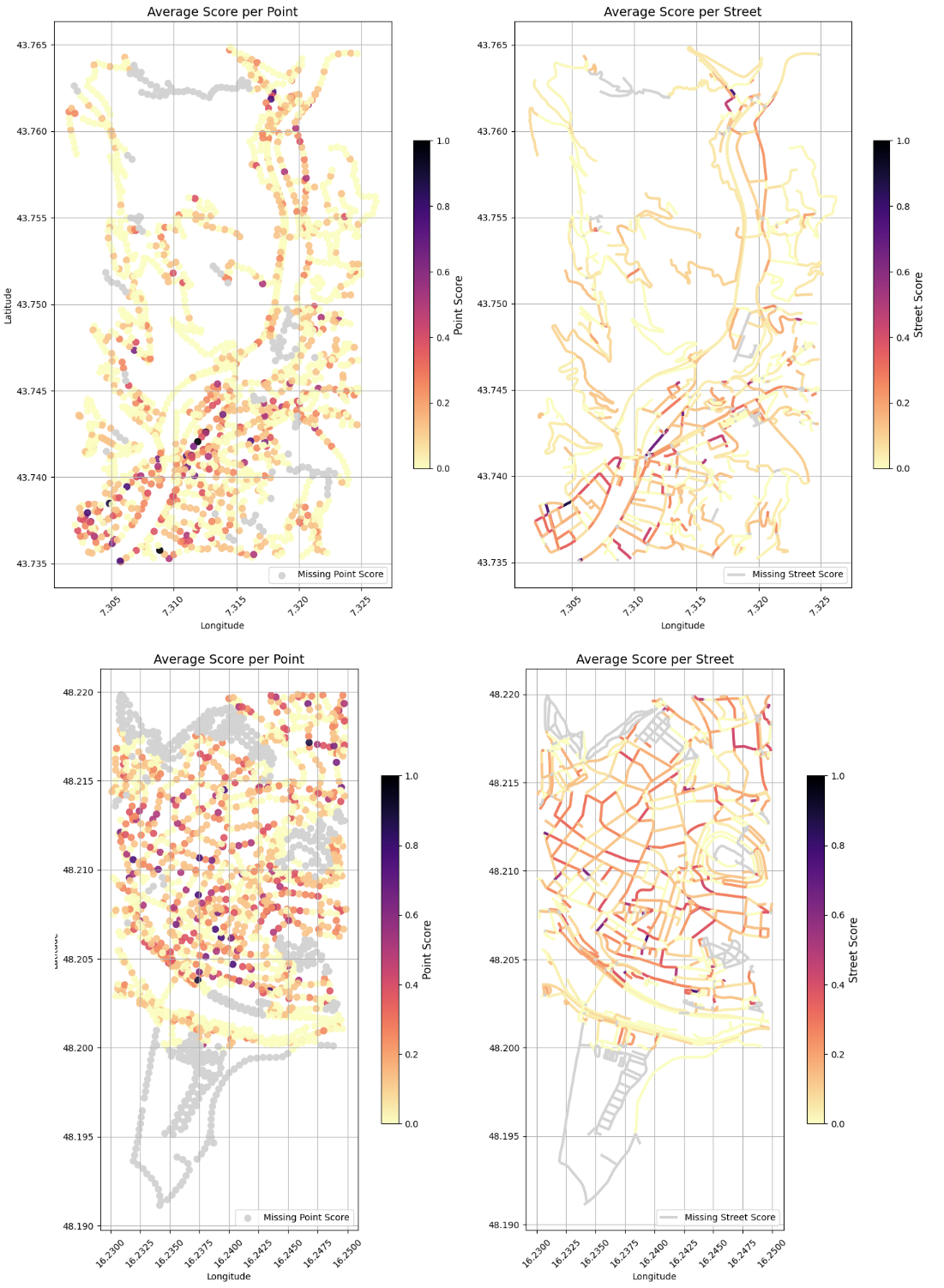}
  \caption{Averaged predicted scores for sidewalk width at the point level (left subpanels) and aggregated/averaged at the street segment level (right subpanels) from Module 4 – Task 3 (Measuring) shown for Nice (top) and Vienna (bottom)}
  \label{fig6}
\end{figure}
Figure ~\ref{fig5} presents the summed outputs for Task 2 (Storefront Counting), visualized as total scores at both the point level and street-segment level (right subpanels) for Nice and Vienna. The underlying hypothesis for the sum statistics is that the 40 m spacing among the sample points is able to produce a good coverage of street segments while minimizing double counting of storefronts. In Nice, the model accurately highlights commercial corridors such as boulevard de l’Ariane and the central axis of Paillon Valley, the Auchan Shopping Mall in the east and the central street of the municipality of Drap in the north, —consistent with both observed commercial density and the high validation accuracy achieved in this case. Conversely, predicted scores drop sharply in low-density or peri-urban areas, aligning with expectations. In Vienna, although several high-score clusters align with known commercial zones—particularly in the southern sector and near key junctions—interpretation is less straightforward due to lower model accuracy in this area, especially for intermediate storefront counts, as confirmed by the manual validation results.

Figure ~\ref{fig6} presents the average predicted sidewalk width scores from Task 3 (Measuring), visualized at both the point level (left subpanels) and as segment-level averages (right subpanels) for the Nice (top) and Vienna (bottom) study areas. Some spatial coherence can be observed in these results: for instance, wider thoroughfares such as Boulevard de l'Ariane in Nice tend to display higher predicted scores than adjacent narrower streets, which aligns with real-world expectations. However, other inconsistencies appear, including scattered high values in locations unlikely to feature wide sidewalks and abrupt variations between adjacent segments. In addition to the averaging of four separate images per location—which tends to dampen values—these patterns reflect the broader range of error types previously discussed for this task. As a result, predicted widths rarely exceed 1 meter, even in visibly wider areas. While some encouraging spatial signals emerge, the overall uncertainty and variability of Task 3 predictions limit our ability to confidently interpret or comment on these results.
\section{Discussion and Perspectives}
\label{sec:6}
To ensure accessibility and broad usability, the current implementation of SAGAI (v1.0) deliberately adopts the lightest possible configuration. The workflow uses the official LLaVA v1.6 Mistral-7B checkpoint in 4-bit quantized format, allowing the full pipeline to run on free-tier environments in Google Colab. This design choice ensures that SAGAI is readily deployable by users without access to high-performance computing resources. Despite this minimal setup, the algorithm already demonstrates strong performance across some visual scoring tasks. Moreover, there is considerable potential to improve accuracy through lightweight strategies alone. For instance, prompt engineering—e.g. instructing the model to return “NA” when uncertain or to append qualifiers like “(unsure)”—can improve robustness without loading a heavier model. Another strategy would be multi-pass voting, where the model is run multiple times on the same image and only the majority result is kept. This helps retain consistent predictions and filter out unstable outputs. Rule-based validation filters, such as suppressing values below or above realistic thresholds or filtering predictions on visually degraded images, may further enhance output quality. These strategies offer a pragmatic way to improve performance while preserving SAGAI’s lightweight and accessible nature.

From a research perspective, multiple extensions are planned to improve the accuracy and flexibility of SAGAI. The current version relies on Mistral-7B, a 7-billion-parameter language model, which balances accuracy and computational cost. Future versions may explore larger or more advanced backbones, including LLaMA-2 13B \cite{llava2023} or Mixtral 8x7B \cite{mixtral2024}, which may capture finer visual nuances and improve interpretability in complex scenes. In parallel, experiments with higher-precision quantization formats (e.g., 8-bit or 16-bit) and newer LLaVA variants \cite{llavaplus2023} (e.g., LLaVA-Plus, LLaVA-NeXT) will be explored, as these already integrate more advanced vision encoders beyond CLIP. These models may offer improved robustness in occluded, cluttered, or ambiguous urban scenes, while preserving computational efficiency. Proprietary multimodal systems such as Gemini or GPT-4V may also be evaluated in order to assess how commercial vision-language architectures compare to LLaVA in the context of SAGAI.

Importantly, the current implementation of SAGAI relies entirely on zero-shot learning: the vision-language model generates structured numerical outputs directly from natural language prompts, without any task-specific fine-tuning. This zero-shot generalization capability is one of the workflow’s core strengths, enabling immediate deployment across different cities or scoring schemes, as long as prompts are clearly defined. However, this flexibility comes with limitations, particularly in visually ambiguous or culturally variable environments. To address this, future releases of SAGAI will include an optional few-shot learning module. This module will allow users to provide small labeled datasets—comprising only a few annotated examples—to locally calibrate the model. Such integration will make it possible to bridge generic prompting and task-specific customization, enabling semi-supervised workflows that improve both precision and reliability in urban visual scoring tasks.

In addition to accuracy, on a standard Google Colab session (free tier), each case study—Nice and Vienna—required approximately 9,000 seconds (2.5 hours) to process the full image set through the LLaVA model. This corresponds to an average throughput of around 1,200–1,300 images per hour, including image loading, preprocessing, and inference. The Street View Batch Downloader (Module 2) required approximately 80 minutes to retrieve and save all images for a single case study (7,000–8,000 images), depending on connection speed and server response times. While these figures already demonstrate reasonable throughput in a cloud-based environment, further acceleration is possible. Future versions may parallelize image downloads and model inference—such as issuing concurrent API requests or batching multiple images for scoring—which would enable SAGAI to scale more efficiently to larger datasets or multi-temporal applications.

Beyond the vision-language component, several additional enhancements could further strengthen the broader SAGAI pipeline. One practical limitation arises from the potential mismatch between the location of the sample points and the actual position from which the Street View images are captured. This discrepancy may introduce spatial noise, especially in dense or topographically complex environments. Future versions of SAGAI could leverage the metadata embedded in Street View imagery—such as precise camera coordinates or heading information—to reassign sample points to the true image capture location. Moreover, the current implementation relies on the Google Street View Static API, which does not support querying historical imagery. However, integrating alternative platforms such as Mapillary or Google’s dynamic API endpoints could eventually enable temporal analyses, allowing researchers to track visual changes in the built environment over time. Finally, instead of capturing four fixed images at orthogonal directions, it may be more efficient to compute the local street orientation and only retrieve views aligned with the primary road axis—i.e., forward and backward along the street segment—thereby reducing redundancy and improving task relevance.

The three tasks implemented within SAGAI—classifying, counting, and measuring—can be flexibly combined to support more specific protocols for streetscape assessment, such as identifying patterns related to streetscapes and public space components within pattern languages (e.g., \cite{alexander1977}, \cite{mehaffy2020},). More broadly, this opens up the possibility of assessing visually observable urban scenes in relation to normative theoretical models. Thus, through the methodological improvements mentioned above—such as the use of heavier AI models, prompt engineering, few-shot learning, and code parallelisation to scale up the spatial extent of urban analysis—SAGAI will open new perspectives for producing automated assessments of visually perceived built environments, integrating fine-grained qualities of the streetscape skin. Diachronic analysis of streetscape transformations over time also offers a promising avenue to assess the impact of underlying urban processes—such as gentrification, pauperization, cultural changes or the effects of specific policies and market trends—since the streetscape skin tends to evolve more rapidly than the skeletal built volumes that support it \cite{dovey2015}.

\section{Conclusion}
\label{sec:7}
This paper introduced SAGAI—Streetscape Analysis with Generative Artificial Intelligence—a lightweight, reproducible workflow that leverages open-access data and generative vision-language models to automate the scoring of street-level urban environments. Building upon recent advances in VLMs such as LLaVA, the pipeline demonstrates that complex visual tasks—including binary classification, object counting, and dimensional estimation—can now be performed consistently and with reasonable accuracy, without task-specific training or heavy computational infrastructure.

Empirical results across two distinct urban settings, Nice and Vienna, show that the system achieves high accuracy in scene classification (over 90\%), moderate precision in storefront detection, and lower—but still informative—performance in estimating sidewalk width. These outcomes illustrate both the promise and the current limits of zero-shot VLMs in urban contexts. Importantly, many misclassifications arose not from outright model errors but from inherent ambiguities in the visual scenes or insufficiently precise prompt formulations. In several cases, the model’s outputs were closer to plausible approximations than actual errors, suggesting that further gains may be achieved through prompt refinement rather than architectural overhaul.

Designed to run entirely in Python and deployable in free-tier Google Colab environments, SAGAI offers a reproducible and extensible workflow for researchers and practitioners alike. Its main contribution is to enrich streetscape analysis with micro-scale elements that are typically assessed visually, as they are absent from available geodatabases. This is achieved by leveraging online street imagery through generative AI algorithms. Its modular architecture makes it suitable for a range of urban applications, including walkability audits, streetscape benchmarking, infrastructure planning, and spatial equity assessments. The system can be readily adapted to new case studies, tasks, or scoring schemes simply by adjusting the prompts, underscoring its potential for broad reuse across geographic contexts and research questions. Yet, while SAGAI's zero-shot setup ensures wide applicability, localized fine-tuning may be required in settings with atypical streetscapes or culturally specific features.

The full codebase of SAGAI, including setup instructions and prompt templates, is available in the public GitHub repository described in subsection~\ref{sec:github}. Future developments will focus on integrating few-shot tuning modules, supporting alternative imagery sources and vision-language models, and expanding the library of scoring prompts. As such, SAGAI provides both a methodological contribution and a practical foundation for the next generation of geospatial research using generative multimodal AI for streetscape analysis. As generative AI continues to evolve, SAGAI exemplifies a transparent, open-source geospatial workflow that leverages generative AI for applied research and planning.
\section*{Acknowledgments}

This work was supported by the emc2 project under the Driving Urban Transition Partnership, co-funded by ANR (France), FFG (Austria), MUR (Italy), and Vinnova (Sweden) with contributions from the European Commission.

\clearpage
\appendix
\section*{Appendix A. Prompt Templates Used in SAGAI v1.0}
\addcontentsline{toc}{section}{Appendix A. Prompt Templates Used in SAGAI v1.0}
\label{app:prompts}

\noindent\textbf{Task T1 – Categorization: Urban vs. Rural}

\vspace{0.5em}
\fbox{%
\begin{minipage}{0.95\textwidth}
\ttfamily\raggedright
 \textit{\{role\_description\}} = You are an AI assistant designed to analyze street-level images. Your task is to determine whether the environment shown in the image is urban or rural.\\
 \textit{\{theory\_model\}} = Classification Guide:\\
- 0: Rural area — sparse built environment, natural surroundings, few or no buildings.\\
- 1: Urban area — dense built environment, visible infrastructure, buildings.\\
 \textit{\{task\}} = Carefully observe the image and determine whether it depicts a rural or urban environment.\\
Use the classification guide above to assign a score.\\
 Return only the classification (0 or 1). Do not explain your answer or add extra text.\\
 \textit{\{response\_format\}} = Answer format: 0 or 1
\end{minipage}%
}

\vspace{1.5em}
\noindent\textbf{Task T2 – Counting: Visible Shopfronts}

\vspace{0.5em}
\fbox{%
\begin{minipage}{0.95\textwidth}
\ttfamily\raggedright
 \textit{\{role\_description\}} = You are an AI assistant designed to analyze street-level images.\\
Your job is to detect the presence of commercial storefronts, such as shops, restaurants, or businesses.\\

 \textit{\{theory\_model\}} = Scoring Guide:\\
- 0: No visible shops or commercial storefronts.\\
- 1: One visible shop or storefront.\\
- 2: More than one shop or storefront is visible.\\

 \textit{\{task\}} = Look at the image carefully and apply the scoring guide above.\\
Return only the score (0, 1, or 2) based on how many shops are visible.\\
Do not explain your answer or add text. Only output the number.\\
 \textit{\{response\_format\}} = Answer format: 0, 1, or 2
\end{minipage}%
}

\vspace{1.5em}
\noindent\textbf{Task T3 – Measuring: Estimated Sidewalk Width}

\vspace{0.5em}
\fbox{%
\begin{minipage}{0.95\textwidth}
\ttfamily\raggedright
 \textit{\{role\_description\}} = You are an AI assistant designed to analyze street-level images. Your task is to estimate the visible width of a sidewalk.\\

 \textit{\{theory\_model\}} = Scoring Guide:\\
- 0: No visible sidewalk or the sidewalk is not clearly identifiable.\\
- Otherwise: Return the estimated width of the sidewalk in meters, rounded to the nearest 0.5 (e.g., 1.0, 1.5, 2.0, 2.5, 3.0).\\

 \textit{\{task\}} = Look at the image carefully. If a sidewalk is visible, estimate its width in meters.\\
If no sidewalk is visible or it's unclear, return 0.\\
Do not explain your answer or add any text. Only output a single number.\\
 \textit{\{response\_format\}} = Answer format: 0 or a number (e.g., 1.0, 1.5, 2.0, 2.5, 3.0)
\end{minipage}%
}

\clearpage
\section*{Appendix B. LLaVA Predictions and Human Annotations}
\addcontentsline{toc}{section}{Appendix B. LLaVA Predictions and Human Annotations}
\label{app:validation}

\begin{center}
  \includegraphics[width=0.85\textwidth]{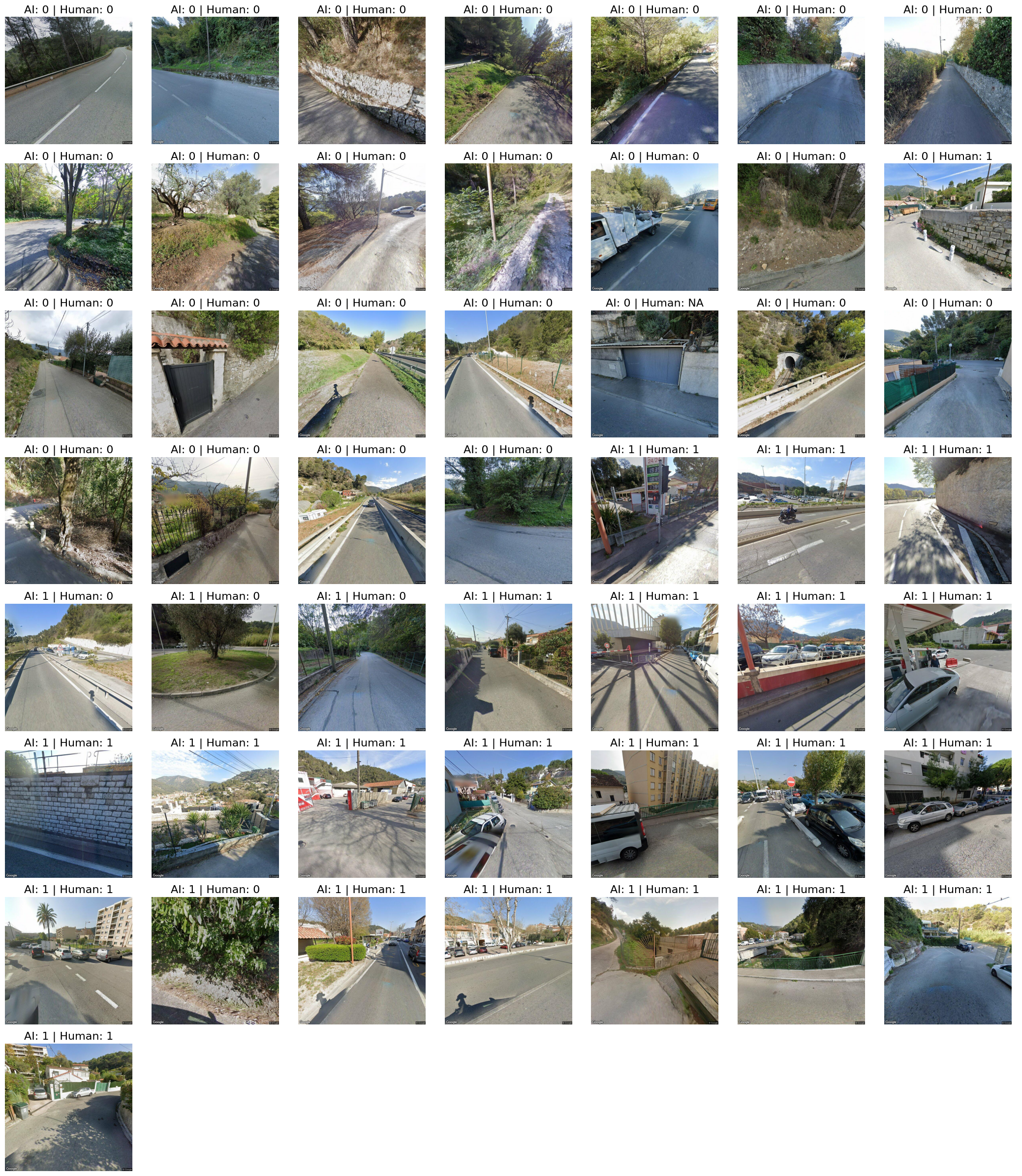}

  \vspace{0.5em}
  \textbf{Task T1 – Nice:} Comparison of LLaVA predictions and human annotations
\end{center}

\clearpage
\begin{center}
  \includegraphics[width=0.85\textwidth]{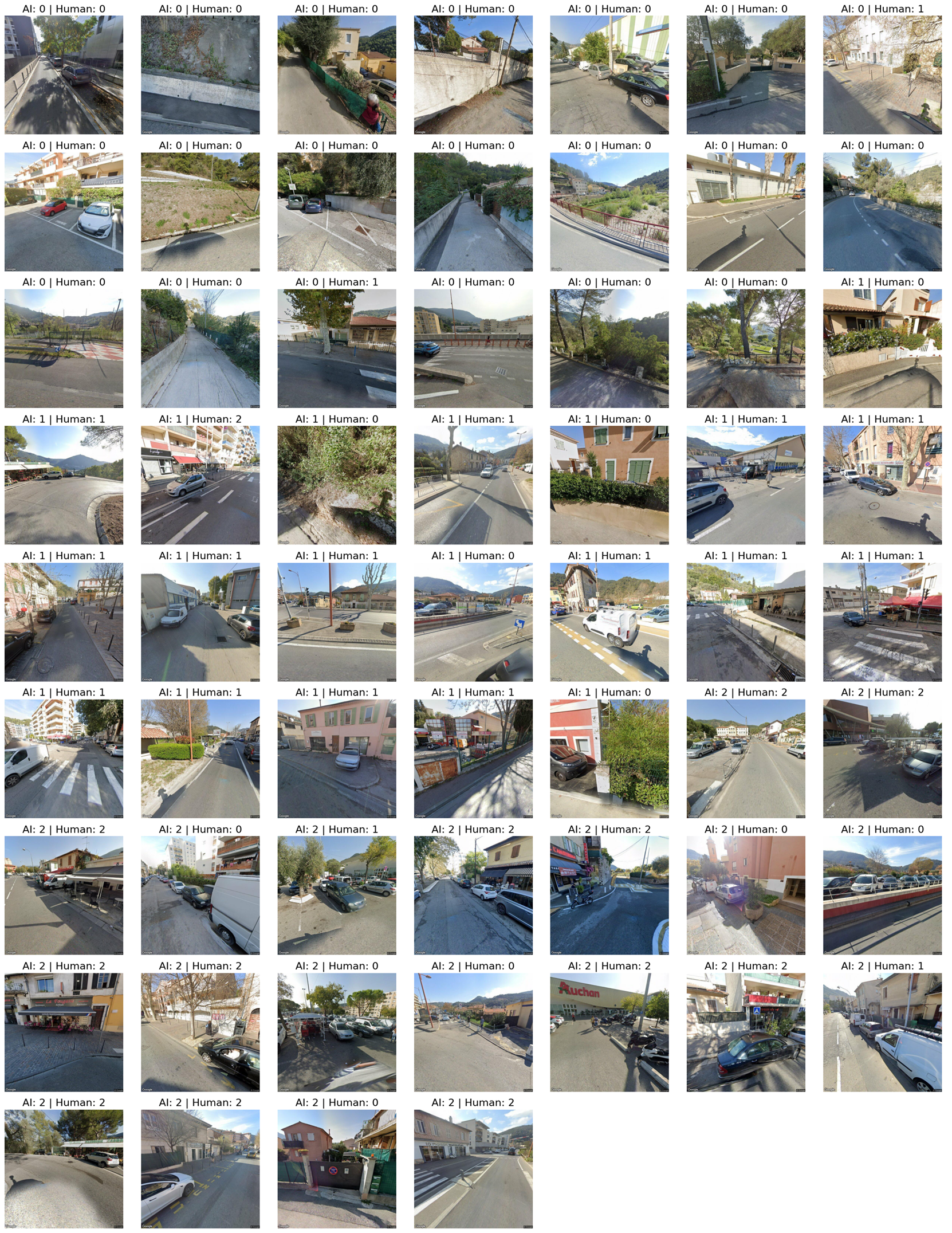}

  \vspace{0.5em}
  \textbf{Task T2 – Nice:} Comparison of LLaVA predictions and human annotations
\end{center}

\clearpage
\begin{center}
  \includegraphics[width=0.85\textwidth]{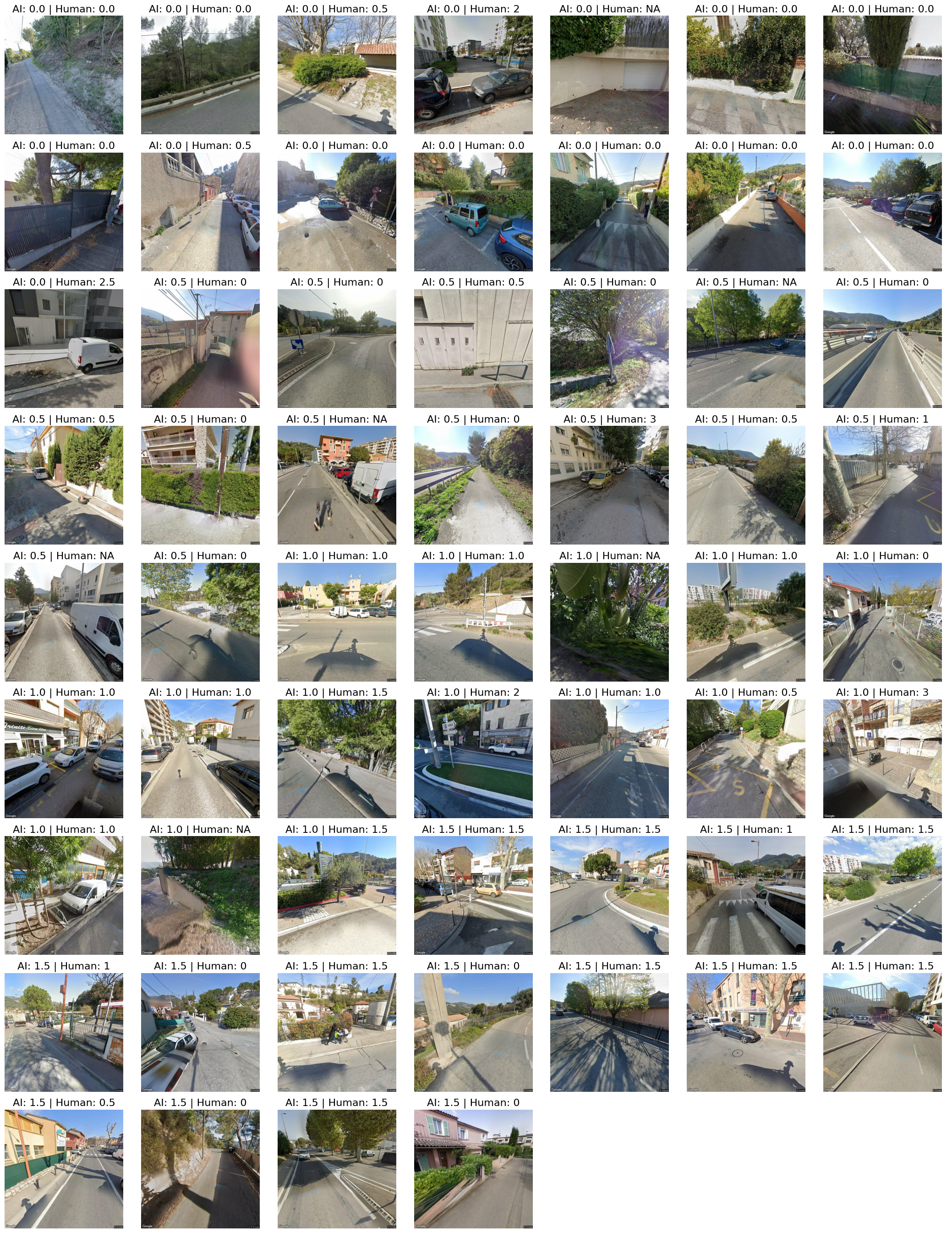}

  \vspace{0.5em}
  \textbf{Task T3 – Nice:} Comparison of LLaVA predictions and human annotations
\end{center}

\clearpage
\begin{center}
  \includegraphics[width=0.85\textwidth]{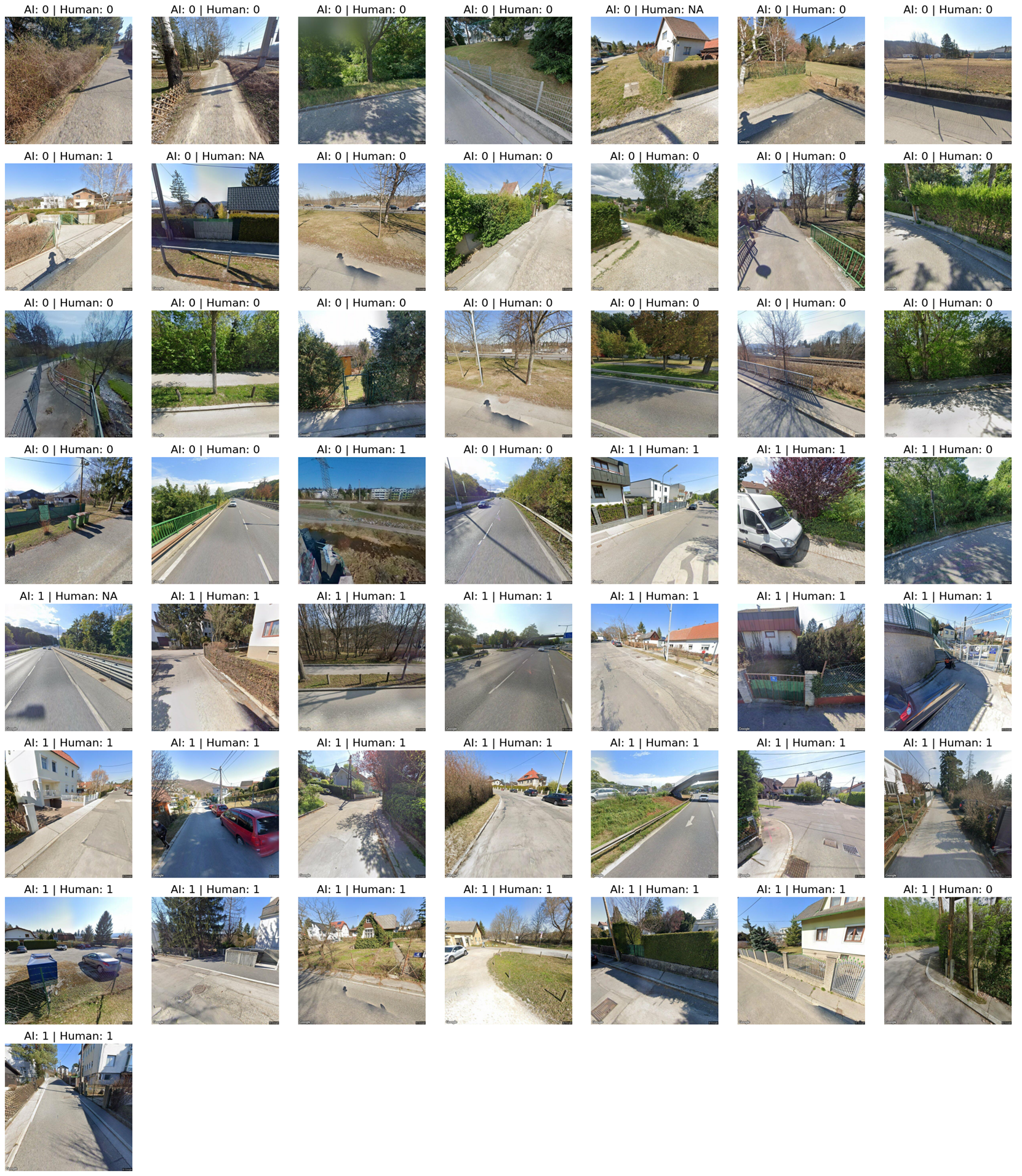}

  \vspace{0.5em}
  \textbf{Task T1 – Vienna:} Comparison of LLaVA predictions and human annotations
\end{center}

\clearpage
\begin{center}
  \includegraphics[width=0.85\textwidth]{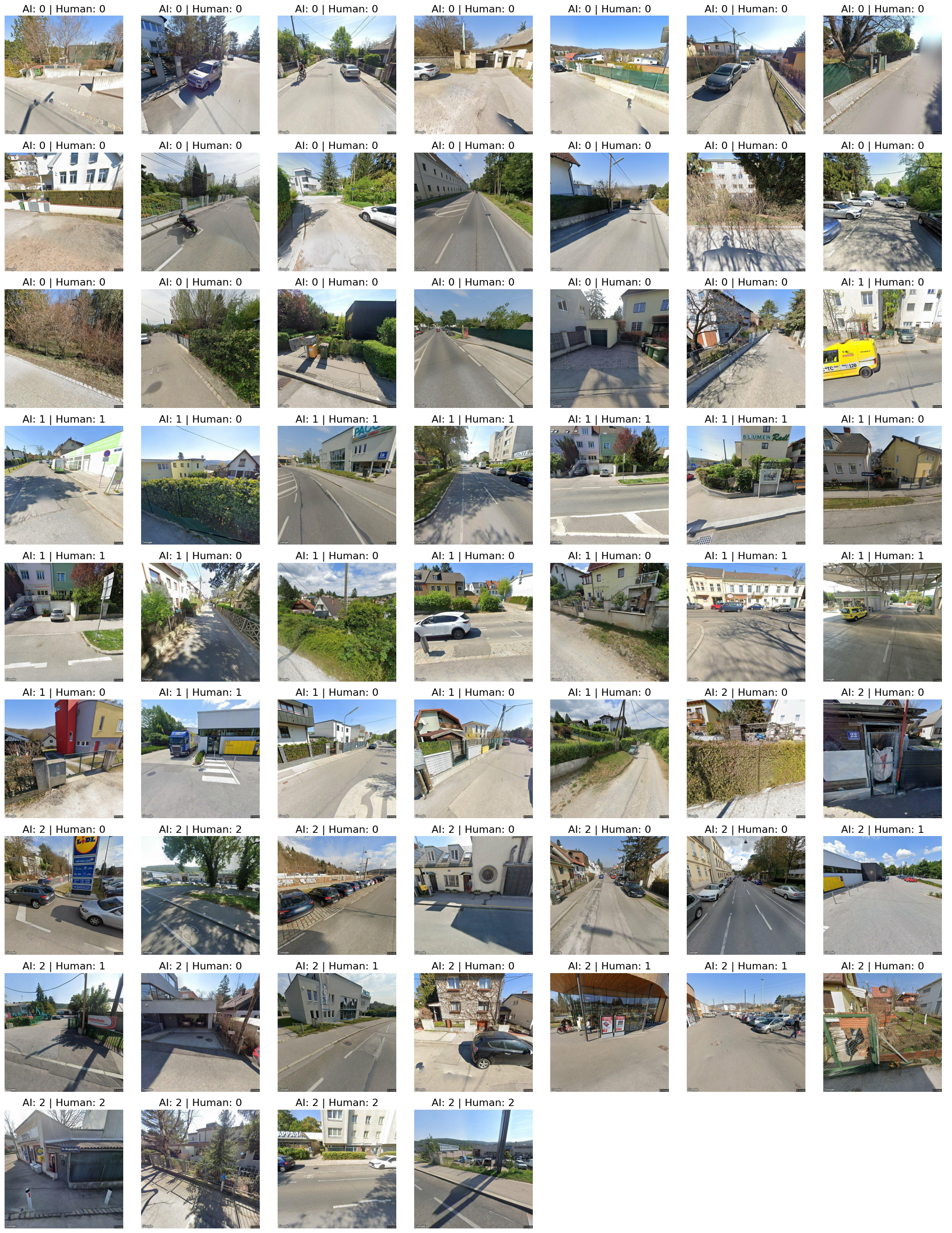}

  \vspace{0.5em}
  \textbf{Task T2 – Vienna:} Comparison of LLaVA predictions and human annotations
\end{center}

\clearpage
\begin{center}
  \includegraphics[width=0.85\textwidth]{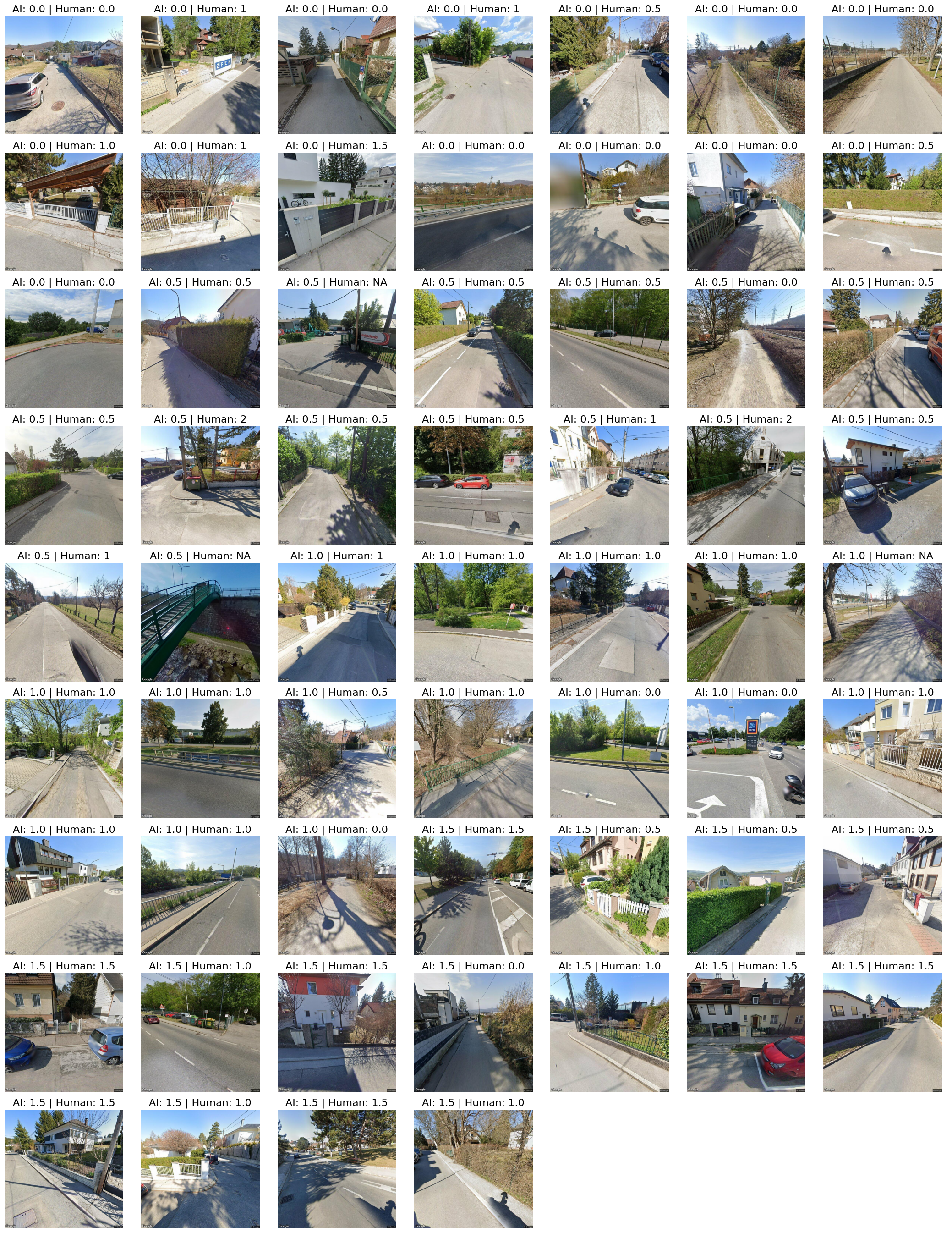}

  \vspace{0.5em}
  \textbf{Task T3 – Vienna:} Comparison of LLaVA predictions and human annotations
\end{center}


\begin{thebibliography}{99}

\bibitem{gehl2013} Gehl, J., Svarre, B. (2013). \textit{How to Study Public Life}. Washington: Island Press. 192 pp.

\bibitem{harvey2016} Harvey, C., Aultman-Hall, L. (2016). Measuring urban streetscapes for livability: A review of approaches. \textit{The Professional Geographer}, 68(1), 149–158.

\bibitem{dovey2018} Dovey, K., Pafka, E., Ristic, M. (2018). \textit{Mapping Urbanities: Morphologies, Flows, Possibilities}. New York: Routledge. 290 pp.

\bibitem{harvey2017} Harvey, C., Aultman-Hall, L., Troy, A., Hurley, S.E. (2017). Streetscape skeleton measurement and classification. \textit{Environment and Planning B: Urban Analytics and City Science}, 44(4), 668–692.

\bibitem{araldi2025} Araldi, A., Fleischmann, M., Fusco, G., Novotný, M. (2025). Streetscape Morphometrics: Expanding Momepy to Analyze Urban Form from the Street Point of View. \textit{SSRN}, \url{http://dx.doi.org/10.2139/ssrn.5051213}

\bibitem{clarke2011} Clarke, P., Ailshire, J., Melendez, R., Bader, M., Morenoff, J. (2011). Using Google Earth to conduct a neighborhood audit: Reliability of a virtual audit instrument. \textit{Health Place}, 16(6), 1224–1229.

\bibitem{rundle2011} Rundle, A., Bader, M., Richards, C., Neckerman, K., Teitler, J. (2011). Using Google Street View to audit neighborhood environments. \textit{American Journal of Preventive Medicine}, 40(1), 94–100.

\bibitem{li2018} Li, X., Ratti, C. (2018). Mapping the spatial distribution of shade provision of street trees in Boston using Google Street View panoramas. \textit{Urban Forestry and Urban Greening}, 31, 109–119.

\bibitem{hosseini2022} Hosseini, M., Miranda, F., Lin, J., Silva, C.T. (2022). CitySurfaces: City-scale semantic segmentation of sidewalk materials. \textit{Sustainable Cities and Society}, 79, 103630.

\bibitem{seiferling2017} Seiferling, I., Naik, N., Ratti, C., Proulx, R. (2017). Green streets: Quantifying and mapping urban trees with street-level imagery and computer vision. \textit{Landscape and Urban Planning}, 165, 93–101.

\bibitem{law2018} Law, S., Seresinhe, C.I., Shen, Y., Gutierrez-Roig, M. (2018). Street-Frontage-Net: Urban image classification using deep convolutional neural networks. \textit{International Journal of Geographical Information Science}, 34(4), 681–707.

\bibitem{gebru2017} Gebru, T., Krause, J., Wang, Y., Chen, D., Deng, J., Aiden, E.L., Fei-Fei, L. (2017). Using deep learning and Google Street View to estimate the demographic makeup of neighborhoods across the United States. \textit{PNAS}, 114(50), 13108–13113.

\bibitem{zhang2024} Zhang, F., Salazar-Miranda, A., Duarte, F., Vale, L., Hack, G., Chen, M., Mantaras, D.A., de Souza, F., Ratti, C. (2024). Urban Visual Intelligence: Studying Cities with Artificial Intelligence and Street-Level Imagery. \textit{Annals of the AAG}, 114(4), 876–897.

\bibitem{girshick2014} Girshick, R., Donahue, J., Darrell, T., Malik, J. (2014). Rich feature hierarchies for accurate object detection and semantic segmentation. In \textit{CVPR}, 580–587.

\bibitem{redmon2016} Redmon, J., Divvala, S., Girshick, R., Farhadi, A. (2016). You Only Look Once: Unified, Real-Time Object Detection. In \textit{CVPR}, 779–788.

\bibitem{wang2024} Wang, S. et al. (2024). Mapping the landscape and roadmap of geospatial AI in quantitative human geography. \textit{IJAEOG}, 128, 103734.

\bibitem{song2023} Song, Y. et al. (2023). Advances in GeoAI for mapping. \textit{IJAEOG}, 120, 103300.

\bibitem{liu2023} Liu, H., Zhang, Y., Xu, Y., Zhang, Z., Yang, Z., Dai, J., Xu, Y. (2023). Visual instruction tuning. \textit{arXiv preprint}, arXiv:2304.08485.

\bibitem{li2023} Li, J. et al. (2023). BLIP-2: Bootstrapping language-image pre-training. arXiv:2301.12597.

\bibitem{openai2023} OpenAI (2023). GPT-4 Technical Report. \url{https://openai.com/research/gpt-4}

\bibitem{alayrac2022} Alayrac, J.B., Donahue, J., Luc, P., Miech, A., Barr, I., Hasson, Y., Momeni, F., Milani, S., Reynolds, M., Borgeaud, S., Azab, M., Smaira, L., Tschannen, M., El-Nouby, A., Gesmundo, A., Ring, R., Rutherford, E., Mensch, A., Setty, R., Brock, A., Botvinick, M., Vinyals, O., Zisserman, A., Simonyan, K., Carreira, J. (2022). Flamingo: a Visual Language Model for Few-Shot Learning. \textit{arXiv preprint}, arXiv:2204.14198.

\bibitem{radford2021} Radford, A., Kim, J.W., Hallacy, C., Ramesh, A., Goh, G., Agarwal, S., Sastry, G., Askell, A., Mishkin, P., Clark, J., Krueger, G., Sutskever, I. (2021). Learning transferable visual models from natural language supervision. In \textit{ICML}, 8748–8763.

\bibitem{dosovitskiy2020} Dosovitskiy, A., Beyer, L., Kolesnikov, A., Weissenborn, D., Zhai, X., Unterthiner, T., Dehghani, M., Minderer, M., Heigold, G., Gelly, S., Uszkoreit, J., Houlsby, N. (2020). An image is worth 16x16 words: Transformers for image recognition at scale. \textit{arXiv preprint}, arXiv:2010.11929.

\bibitem{liu2021} Liu, Z., Lin, Y., Cao, Y., Hu, H., Wei, Y., Zhang, Z., Lin, S., Guo, B. (2021). Swin Transformer: Hierarchical vision transformer using shifted windows. In \textit{ICCV}, 10012–10022.

\bibitem{li2024} Li, Z. et al. (2024). StreetViewLLM. arXiv:2411.14476.

\bibitem{duan2024} Duan, Z. et al. (2024). CityLLaVA. In \textit{CVPR Workshops}, 7180–7189.

\bibitem{blecic2024} Blečić, I., Saiu, V., Trunfio, G.A. (2024). Enhancing Urban Walkability Assessment with Multimodal LLMs. In \textit{ICCSA 2024 Workshops}, LNCS 14819, 394–411.

\bibitem{wei2024} Wei, C. et al. (2024). GeoTool-GPT. \textit{IJGIS}, 39(4), 707–731.

\bibitem{kazemi2024} Kazemi Beydokhti, M. et al. (2024). Probabilistic qualitative spatial reasoning. \textit{IJGIS}, 39(4), 817–846.

\bibitem{schmidt2025} Schmidt, S. et al. (2025). Assessing spatial accuracy of geocoding flood-related imagery. \textit{Spatial Information Research}, 33(15).

\bibitem{fusco2024} Fusco, G., Berghauser Pont, M., Cutini, V., Psenner, A. (2024). Guiding principles for the 15-minute city. In \textit{AESOP 2024 Proceedings}, 690–707. \url{https://hal.science/hal-04798781}

\bibitem{fusco2025} Fusco, G., Picard, G. (2025). Assessing Form Patterns for the Suburban 15-Minute City: The Case of Drap (France). In \textit{ISUF 2025 Proceedings}, in press.

\bibitem{alexander1977} Alexander, C., Ishikawa, S., Silverstein, M. (1977). \textit{A Pattern Language}. New York: Oxford University Press. 1071 pp.

\bibitem{gehl2010} Gehl, J. (2010). \textit{Cities for People}. Washington: Island Press. 288 pp.

\bibitem{yamada2010} Yamada, I., Thill, J.C. (2010). Local indicators of network-constrained clusters. \textit{Annals of the AAG}, 100(2), 269–285.

\bibitem{llava2023} Liu, H., Li, C., Wu, Q., Lee, Y.J. (2023). LLaVA: Large Language and Vision Assistant. \textit{arXiv preprint} arXiv:2302.13971.

\bibitem{mixtral2024} Jiang, A.Q., Sablayrolles, A., Roux, A., Mensch, A., Savary, B., Bamford, C., Singh Chaplot, D., de las Casas, D., Bou Hanna, E., Bressand, F., Lengyel, G., Bour, G., Lample, G., Renard Lavaud, L., Saulnier, L., Lachaux, M.-A., Stock, P., Subramanian, S., Yang, S., Antoniak, S., Le Scao, T., Gervet, T., Lavril, T., Wang, T., Lacroix, T., El Sayed, W. (2024). Mixtral of Experts. \textit{arXiv preprint}, arXiv:2401.04088.

\bibitem{llavaplus2023} Liu, S., Cheng, H., Liu, H., Zhang, H., Li, F., Ren, T., Zou, X., Yang, J., Su, H., Zhu, J., Zhang, L., Gao, J. (2023). LLaVA-Plus: Learning to Use Tools for Creating Multimodal Agents. \textit{arXiv preprint}, arXiv:2311.05437.

\bibitem{mehaffy2020} Mehaffy, M., Kryasheva, Y., Rudd, A., Salingaros, N. (2020). \textit{A New Pattern Language for Growing Regions: Places, Networks, Processes.}. Portland (OR): Sustasis Press.

\bibitem{dovey2015} Dovey, K., Wood, S. (2015). Public/private urban interfaces: Type, adaptation, assemblage. \textit{Journal of Urbanism}, 8(1), 1–16.

\end{thebibliography}
\end{document}